\pdfoutput=1
\documentclass[lettersize, journal]{IEEEtran}

\newif\ifarxiv
\arxivtrue %

\usepackage[numbers]{natbib}

\usepackage{authblk}

\usepackage{blindtext, verbatim}

\usepackage{amsmath,amsfonts,amsthm, amssymb, bm,stmaryrd} 
 
\usepackage{graphicx, subcaption}

\usepackage{color}

\usepackage{hyphenat}

\usepackage{url}

\usepackage{breakurl}
\usepackage[breaklinks]{hyperref}

\newcommand{\boxpaper}[1]{}
\newcommand{\bluecolor}[1]{\textcolor{black}{#1}}

\newcommand{\bluecolorr}[1]{\textcolor{black}{#1}}

\ifarxiv
\usepackage{balance}
\usepackage[absolute]{textpos}
\fi

\title{Quantifying Noise of Dynamic Vision Sensor}

\author[1]{Evgeny V.~Votyakov}
\author[1]{Alessandro Artusi}
\affil[1]{DeepCamera MRG, CYENS Centre of Excellence, Nicosia, Cyprus}

\begin{document}
\maketitle

\ifarxiv
\definecolor{somegray}{gray}{0.6}
\newcommand{\darkgrayed}[1]{\textcolor{somegray}{#1}}
\begin{textblock}{11}(2.5, 0.4)
\begin{center}
\darkgrayed{This work has been submitted to the  IEEE Signal Processing Letters. Copyright may be transferred without notice, after which this version may no longer be accessible.
\copyright IEEE}
\end{center}
\end{textblock}
\fi

\begin{abstract}

Dynamic Vision Sensors (DVS) exhibit significant background activity (BA) noise that is mixed with the underlying sensor signal. Due to the dynamic nature of the data and the absence of ground truth in practical scenarios, separating noise from signal using conventional image-processing methods is challenging.
This letter presents a novel approach for characterizing BA noise based on Detrended Fluctuation Analysis (DFA). The proposed method enables quantitative characterization of noise and signal without ground truth and provides a principled way to derive optimal denoising filter parameters. Its effectiveness is demonstrated on widely used dataset in the field.
\end{abstract}



\section{Introduction}

Dynamic Vision Sensors (DVS), also known as silicon retinas, are event-based vision sensors that asynchronously detect per-pixel changes in brightness \cite{DVSsurvey2020}. A major limitation of DVS is the presence of significant background activity (BA) noise, which is intermixed with the useful signal and can severely limit performance in practical applications. Although careful sensor parameters tuning can reduce BA noise \cite{Gallego2019FocusIA,Muglikar2021HowTC,Shiba2022SecretsOE,Graa2023OptimalBA,Graa2023ShiningLO}, large portions of the event stream may still be corrupted, as BA noise is difficult to reliably identify and remove.

\bluecolor{State-of-the-art DVS denoising methods are largely based on the assumption that noise events are spatially and temporally isolated. This principle, introduced by Delbrück \cite{Delbruck2008FramefreeDD,Liu2015DesignOA}, relies on spatiotemporal neighborhood checks within a time window $\Delta T$. Several extensions have been proposed, including faster neighbor searches (e.g., k-noise \cite{khodamoradi2018n}, DWF \cite{guo2022low}), expanded 3D filtering criteria \cite{feng2020event}, and weighted spatiotemporal projections \cite{lagorce2016hots,IETSfilter,guo2022low}. More recently, deep-learning-based approaches have also been explored \cite{9091226,baldwin2020eventEDnCNN,Fang2022AEDNet}.}

Assessing the quality of denoised DVS data remains challenging. First, results strongly depend on the event accumulation time: short intervals yield sparse data, while long intervals may blur the signal and cause misclassification as noise. Second, BA filters require careful parameter tuning, and optimal values vary with the accumulation time. Objective evaluation is therefore difficult. Ground-truth-based evaluation is limited to synthetic datasets \cite{guo2022low}, while application-driven evaluations, such as warped-event contrast maximization \cite{Gallego2018}, are computationally complex. Event-wise noise labeling has also been proposed \cite{baldwin2020eventEDnCNN,Fang2022AEDNet}, but is impractical due to the high data rates of DVS \cite{Xu2023DenoisingFor}.

\bluecolor{These limitations motivate an alternative perspective that avoids classifying individual events as noise or signal. In this letter, we introduce a novel objective criterion for evaluating DVS denoising performance based on the statistical properties of event time series. Specifically, we quantify BA noise using detrended fluctuation analysis (DFA). The goal is to assess the quality of the filtered signal and residual noise, rather than explicitly separating noise and signal at the event level. The resulting noise metric enables principled optimization of denoising filter parameters.}

After briefly reviewing DFA, we adapt it to DVS data and demonstrate its effectiveness on a widely used datasets in the field, where it is employed to identify optimal denoising filter settings. To the best of our knowledge, this is the first application of DFA to the analysis of DVS data.

\section{Dynamic Vision Sensor (DVS)} 

Dynamic Vision Sensors (DVS) \cite{DVSsurvey2020} is a 2D matrix of  asynchronously pixels.  Each pixel has  static and dynamic variables. The static one is for  previous reference illumination of the pixel and the dynamic one is for the  instantaneous illumination. At every tact, the instantaneous and reference illuminations are compared. If their difference is more than a threshold, the pixel updates its reference and triggers an event $e=e(t, x, y, p)$, being $t$  an instance of time, when the event is triggered, $(x,y)$ pixel's 2D coordinates, and $p=\pm 1$ is polarity of the triggered event (positive, if illumination increases, and negative otherwise). As a result, DVS data is a stream of events $\{e_1, e_2, \dots, e_N\}$.

\section{Background Activity (BA) filter}

A simple BA filter works as follows \cite{Delbruck2008FramefreeDD}, \cite{Liu2015DesignOA}, \cite{khodamoradi2018n}, \cite{Yang2018TheSC}, \cite{guo2022low}. Let $\sigma_{ij}$ be a Boolean variable for  a  correlation between $e_i$ and $e_j$  events  in the 3D box $(\Delta S_x, \Delta S_y, \Delta T)$:
\begin{align}
\sigma_{ij} &= |x_i-x_j| \le \Delta S_x \nonumber \\ 
& \land  |y_i-y_j| \le \Delta S_y \land |t_i-t_j| \le \Delta T.   
\label{eq:BAFdefintion}
\end{align} 

When $e_j$ is vicinity of $e_i$,  then  $\sigma_{ij}$ is true. With $\llbracket \cdot \rrbracket$ as  Iverson bracket  ($\llbracket \text{true} \rrbracket=1$,$\llbracket \text{false} \rrbracket=0$),  the total number of correlated events $\rho_i$ in $e_i$'s vicinity is given by
\begin{align}
    \rho_i(\Delta S_x, \Delta S_y, \Delta T) &= \sum_{j\neq i} \llbracket \sigma_{ij} \rrbracket, 
\end{align} Then 
\begin{align}
    e_i(t_ix_i,y_i,p_i) = & \begin{cases}
      \text{signal},  \text{ if } \rho_i \geq \rho_\text{min}, \\
      \text{noise, otherwise.}
      \end{cases}      
\end{align} \bluecolor{These equations are for analyzing the signal offline, and they do not consider the events causality.} We used $\Delta S_x = \Delta S_y =1$, $\rho_\text{min}=1$, and $\Delta T$ is  a single optimizing parameter. \bluecolorr{Note that highly localized, repetitive noise sources such as hot-pixels or fixed-pattern noise (FPN) are assumed to be masked out during a standard preprocessing stage prior to filtering, as their deterministic nature would artificially skew the event time-series statistics.}

Now, we take a time series (TS) of DVS events and split it into clean and noise parts with a BA fliter,  Fig.~\ref{fig:principle}a. The clean signal, $(TS)_{clean}$, is composed of correlated events, i.e.  it  contains long term correlations as much as possible. The noise signal, $(TS)_{noise}$, must be as random as possible, i.e. it contains long term correlations as less as possible.

\section{Detrended Fluctuation Analysis} 

 Detrended Fluctuation Analysis (DFA) \cite{Peng1994} is a statistical method  to discover long-term correlations in a time series. DFA is  numerously exemplified: anaerobic threshold derived from heart rate variabilities \cite{Rogers2020-jn}, earthquakes   \cite{Kataoka2021},  lightning thunderstorms flash sequences in  atmosphere science \cite{Gou2018lightningflash}, financial markets  \cite{Shrestha2021finance}. In image processing, Ramirez \cite{Ramirez2005} explored  DFA at describing roughness of images. Technical aspects are given by Hu \cite{Hu2001} and Kantelhardt \cite{Kantelhardt2001}. For DVS applications, the DFA has  not been yet applied. 

\begin{figure}
\centering
\includegraphics[width=0.49\textwidth]{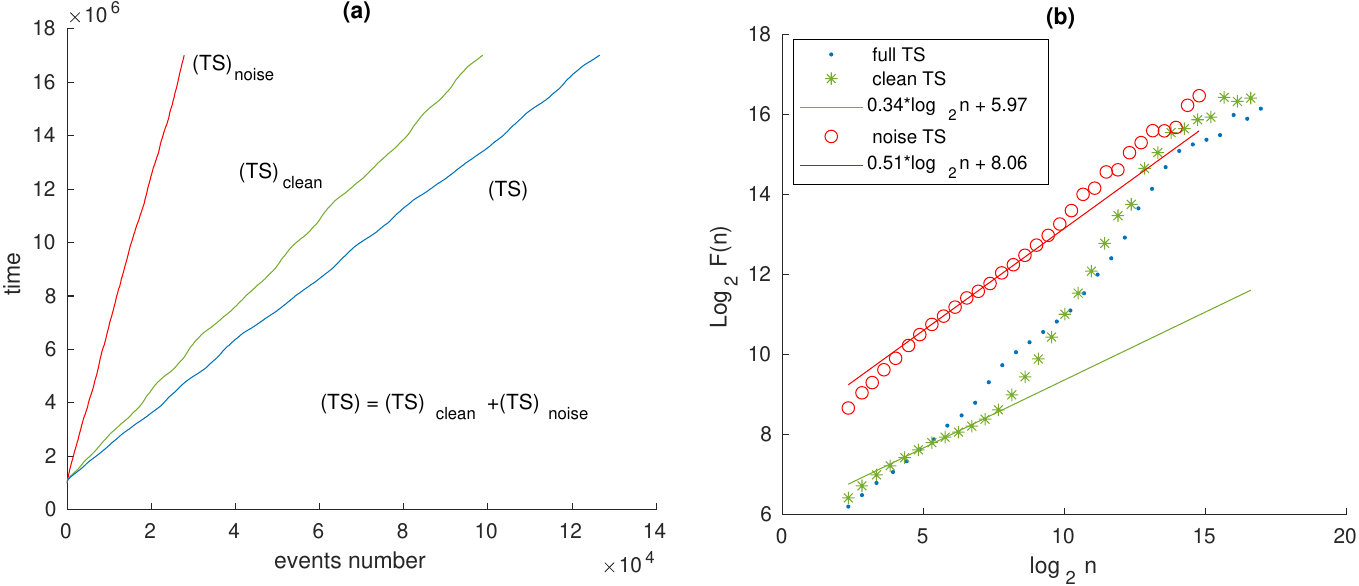} 
\caption{(a) clean $(TS)_\text{clean}$ and noise $(TS)_\text{noise}$  are produced by a BA filter (\ref{eq:BAFdefintion}) from an initial $(TS)_\text{full}$. (b) DFA results. Slope of the line is DFA scaling exponent $\alpha$  characterizing an impact of long term correlations.}
\label{fig:principle}
\end{figure}

Let $x$ be a random variable, and there is a finite series composed of $x$:
\begin{align}
    \{ x_1, x_2, \dots, x_N \}.
    \label{eq:initseries}
\end{align} In the case of DVS, there is a time series of events, and as a random variable $x$ stands  a time interval  $\delta t_i=t_i-t_{i-1}$ between adjacent events, \bluecolor{which is convenient because of then DFA cumulative sums are given by $t_i$}.

Let $X$ be a new random variable equal to a cumulative sum, $X_1  = x_1$, $X_2  = x_1 + x_2$, $\dots$, $X_N  = x_1 + x_2 + \dots + x_N$. Notice that for a DVS stream, the  cumulative sums are  exactly timestamps: $X_i =t_i$. Compose a cumulative series:
\begin{align}
    \{ X_1, X_2, \dots, X_N \}.    \label{eq:X} 
\end{align}  Series (\ref{eq:X}) is partitioned in local segments of equal length $n$, and an independent  least-square fit (local trend) is performed inside the $k$-th segment, $1 \le k \le \lfloor N/n \rfloor$.  Let $\hat{X}_{i}$ be the values  predicted by a local trend, then, $(X_i-\hat{X}_{i})$ are detrended fluctuations. Mean \textit{local} fluctuation of the $k$-th segment is:
\begin{align}
    F_k(n) = \left\{ \frac{1}{n}\sum_{l=in+1}^{in+n}\left(X_l-\hat{X}_l\right)^2 \right\}^{1/2},
    \label{eq:DFAlocal}
\end{align} and the mean \textit{total} fluctuation depending on $n$ is averaged over all mean $k$ local fluctuations: 
\begin{align}
    F(n) = \left\{ \frac{1}{\lfloor N/n \rfloor }\sum_{k=1}^{\lfloor N/n \rfloor }F_k(n)^2 \right\}^{1/2}.
    \label{eq:DFAend}
\end{align}

The computation of eq. \ref{eq:DFAlocal} and \ref{eq:DFAend} run for segments of regularly increasing length $n\in \{m_1, m_2, \dots, m_M$\}, where integers $m_k$ must obey the condition $m_k\simeq q m_{k-1}$, $q>1$ is a constant, i.e. $m_k$ must form  a geometric progression, being $m_1=4$ and  $m_M\simeq N$. As a result, a fluctuation function $F(n)$ is produced.  If the series (\ref{eq:initseries}) is of long-range power-law correlations, $F(n)$ increases by a power-law:
\begin{align}
    F(n)\simeq Cn^\alpha. 
\end{align} The  scaling exponent $\alpha$ gives a metric on self-correlations inside the initial series (\ref{eq:initseries}). To find $\alpha$, it is convenient to make double $\log n -\log F(n)$ plot as shown in Fig.~\ref{fig:principle}b: $\alpha\simeq 0.5$ means  uncorrelated data,  $\alpha > 0.5$ positive long-range correlations, $\alpha < 0.5$  - anticorrelation, and  $\alpha > 1$ the data non-stationary.  \bluecolor{DFA and noise distributions (Gaussian vs. Non-Gaussian),  additive noise effects, colored noise and DFA scaling are discussed in \cite{Kantelhardt2001},  \cite{Taqqu1995}, \cite{Hoell2019Theoret}}.

\section{Results}

\subsection{Simple dataset}

\bluecolor{We emphasize that the focus of this paper is the introduction of a novel metric for determining the optimal parameters of a denoising filter without requiring ground truth data—a capability not previously proposed for DVS signals. As no prior work exists for direct comparison, we objectively assess the performance of BA denoising using a standard dataset (the slot-car dataset) as input noise, applying Delbrück’s BA filter \cite{Delbruck2008FramefreeDD}.} Time bins ranging from $4\times10^6$ to $16\times10^6~\mu$s, with a starting time of $10^6~\mu$s, are considered. The time-bin duration does not affect the results, except that longer bins (up to $32\times10^6~\mu$s) improve statistical reliability, while shorter bins are preferable for 3D $xyt$ visualization, as long bins produce cluttered representations.

\begin{figure}
\centering
\begin{subfigure}{0.24\textwidth}
\includegraphics[width=\textwidth]{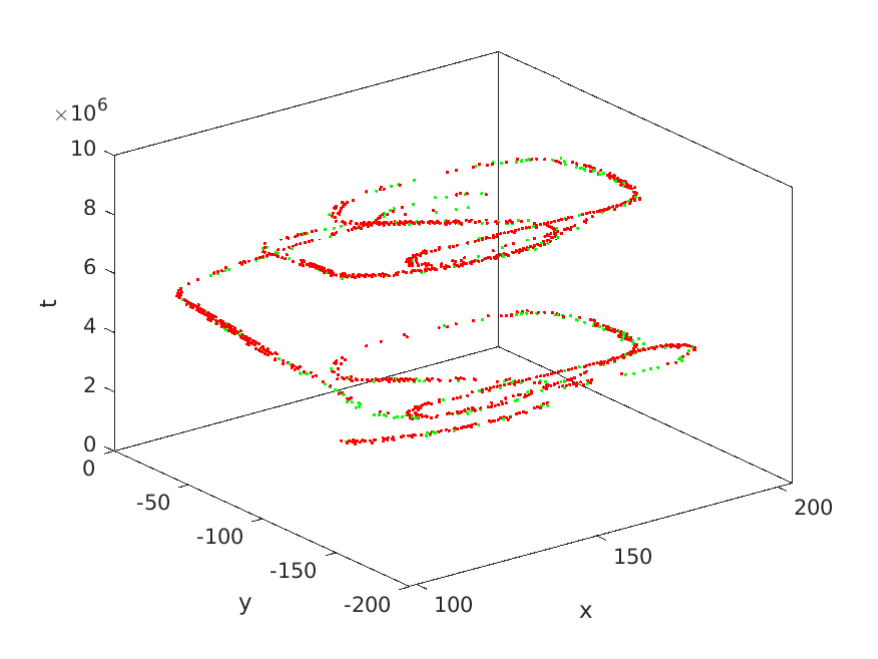} 
\subcaption{$\Delta T= 1000$ clean}
\label{fig:xyt:cleandelT1000}
\end{subfigure}
\begin{subfigure}{0.24\textwidth}
\includegraphics[width=\textwidth]{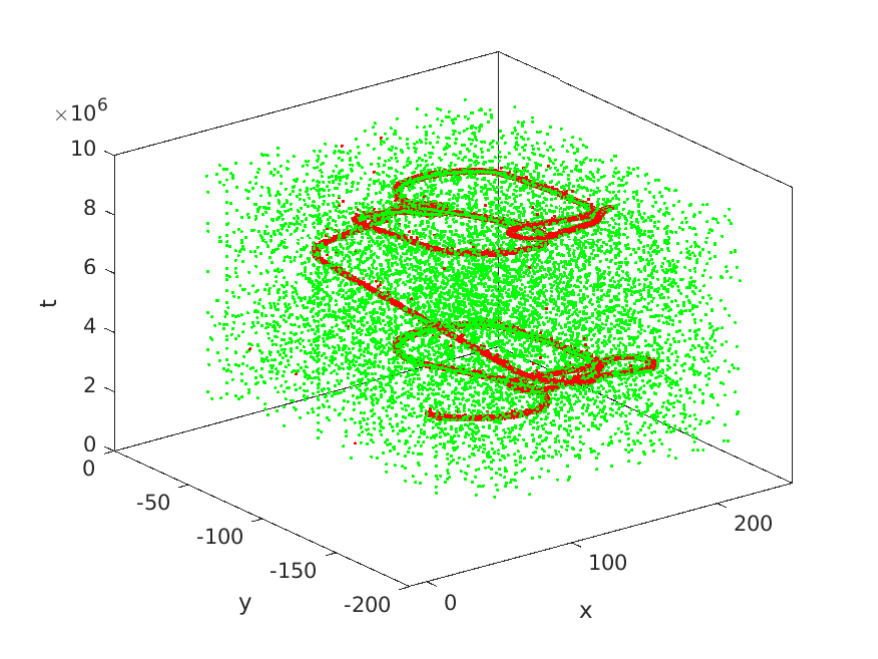}
\subcaption{$\Delta T= 1000$ noise}
\label{fig:xyt:noisdelT1000}
\end{subfigure}
\begin{subfigure}{0.24\textwidth}
\includegraphics[width=\textwidth]{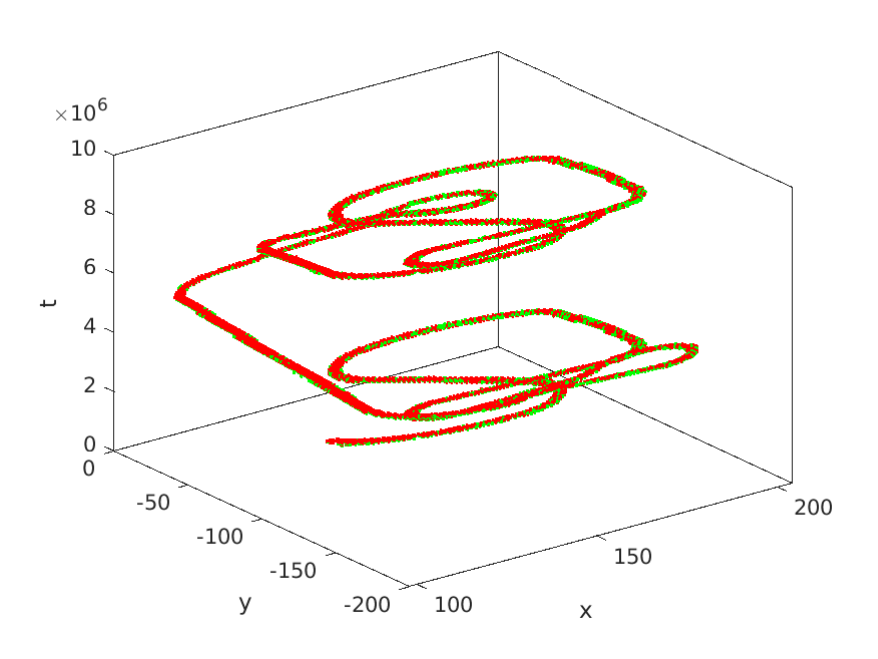} 
\subcaption{$\Delta T= 4000$ clean}
\label{fig:xyt:cleandelT4000}
\end{subfigure}
\begin{subfigure}{0.24\textwidth}
\includegraphics[width=\textwidth]{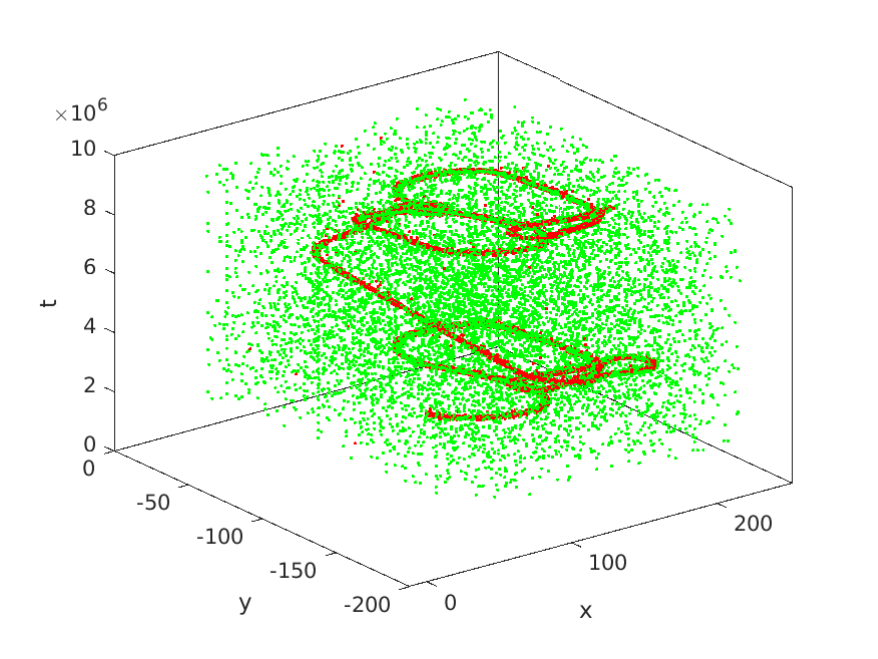}
\subcaption{$\Delta T= 4000$ noise}
\label{fig:xyt:noisdelT4000}
\end{subfigure}
\begin{subfigure}{0.24\textwidth}
\includegraphics[width=\textwidth]{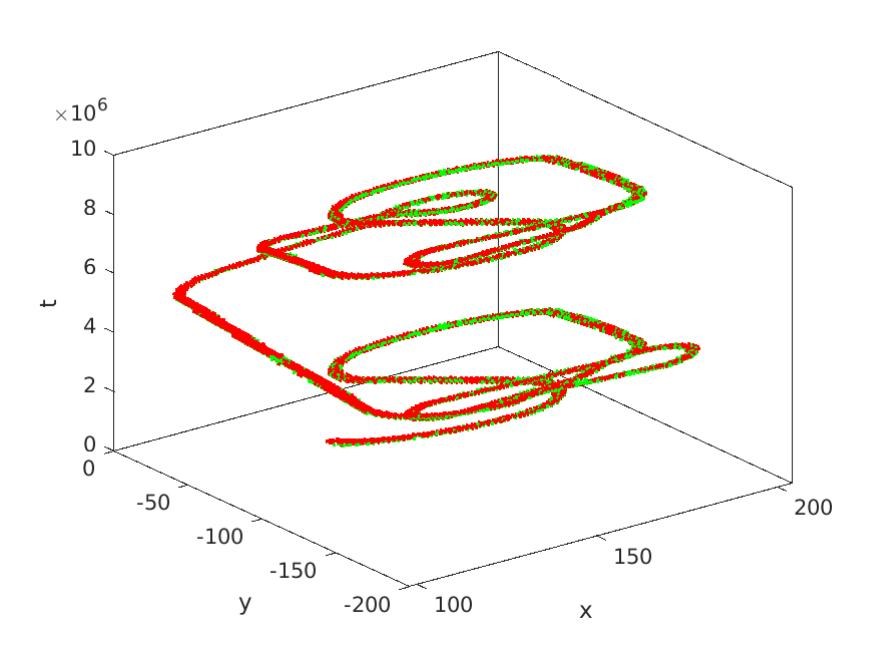} 
\subcaption{$\Delta T= 8000$ clean}
\label{fig:xyt:cleandelT8000}
\end{subfigure}
\begin{subfigure}{0.24\textwidth}
\includegraphics[width=\textwidth]{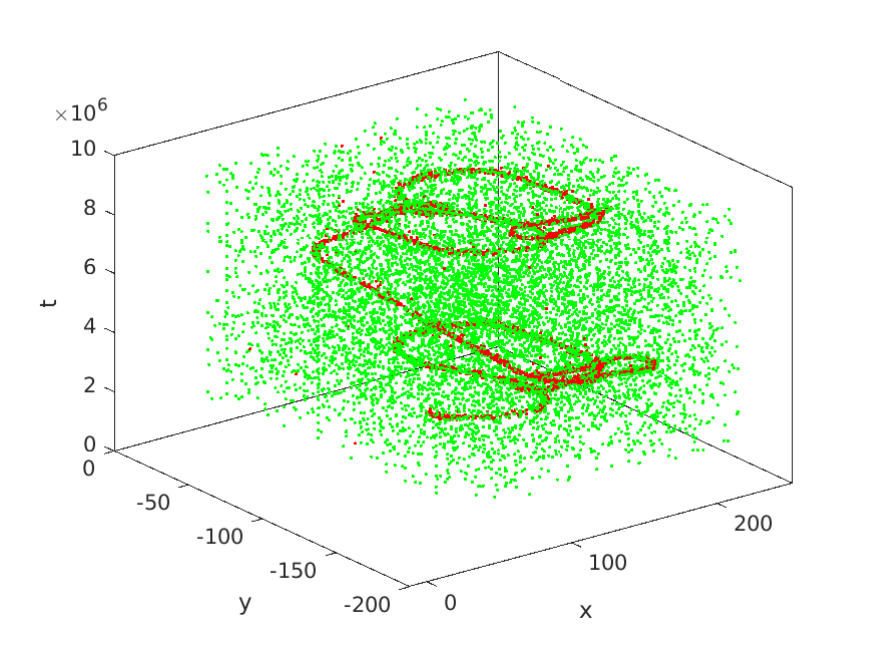}
\subcaption{$\Delta T= 8000$ noise}
\label{fig:xyt:noisdelT8000}
\end{subfigure}
\begin{subfigure}{0.24\textwidth}
\includegraphics[width=\textwidth]{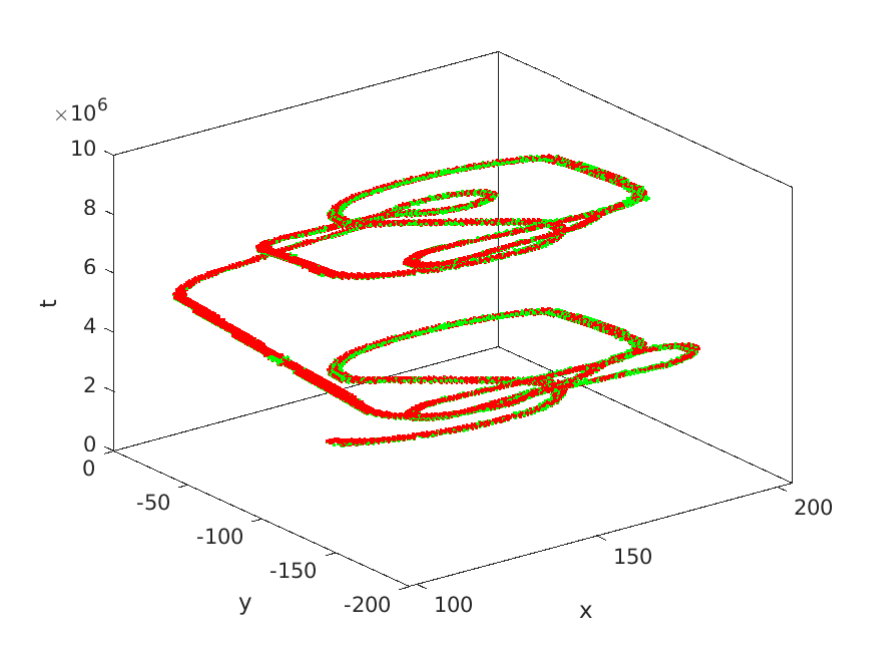} 
\subcaption{$\Delta T= 16000$ clean}
\label{fig:xyt:cleandel168000}
\end{subfigure}
\begin{subfigure}{0.24\textwidth}
\includegraphics[width=\textwidth]{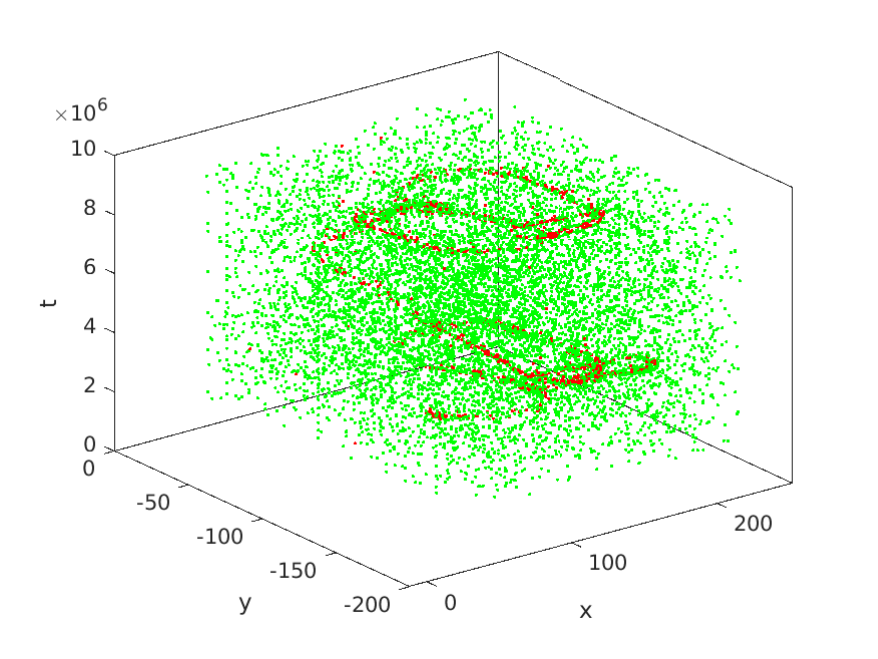}
\subcaption{$\Delta T= 16000$ noise}
\label{fig:xyt:noisdelT16000}
\end{subfigure}
\caption{ BAF applied at various $\Delta T$.}
\end{figure}

Figures~\ref{fig:xyt:cleandelT1000}–\ref{fig:xyt:noisdelT16000} illustrate BA denoising results for $\Delta T$ ranging from 1000 to 16000. While BA noise is effectively removed in all cases, parts of the signal are also suppressed, which is clearly visible in the 3D representations. Due to the simplicity of the dataset, signal traces can still be visually identified in the filtered noise streams.

\begin{figure}
\centering
\includegraphics[width=0.4\textwidth]{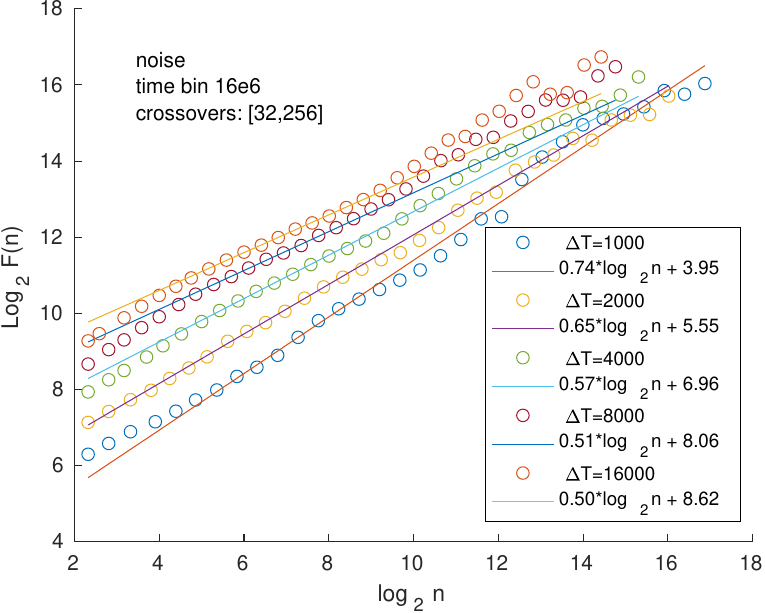}
\caption{DFA applied to the noise time series. }
\label{fig:DFAscaling}
\end{figure}

We now demonstrate how residual signal components captured by the BA filter alter the statistical properties of the noise time series. Figure~\ref{fig:DFAscaling} shows DFA results for the noise streams corresponding to the $\Delta T$ values in Fig.~\ref{fig:xyt:cleandelT1000}–\ref{fig:xyt:noisdelT16000}. The DFA exponent $\alpha$ is obtained from the slope of the scaling regions (see Fig.~\ref{fig:DFAscaling}b). For $\Delta T=1000$, $\alpha=0.74$, indicating correlated noise, whereas for $\Delta T=16000$, $\alpha=0.50$, consistent with uncorrelated (random) noise. This observation agrees with the visual inspection: at small $\Delta T$, spatially ordered signal traces remain in the noise stream, while they are largely absent at larger $\Delta T$.

\begin{figure}[b]
\centering
\includegraphics[width=0.4\textwidth]{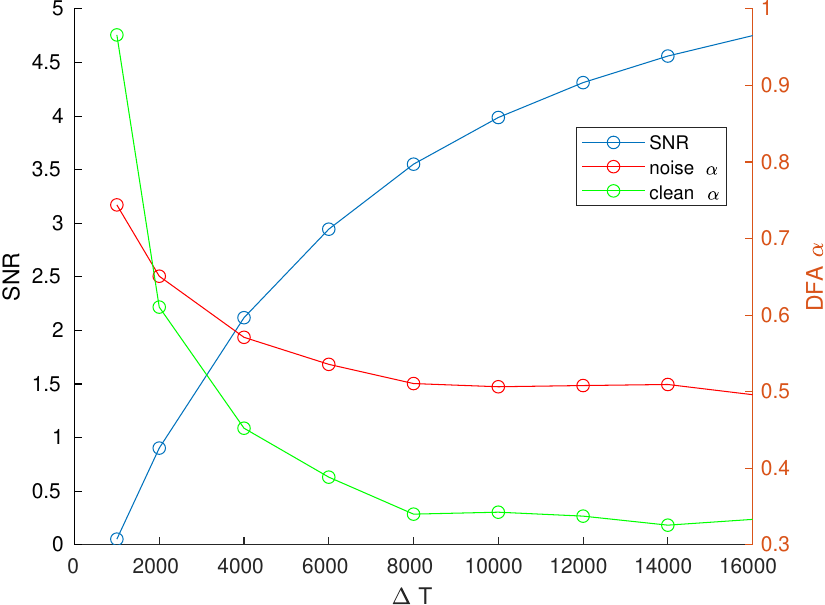} 
\caption{SNR and $\alpha$ as a function of $\Delta T$.}
\label{fig:summary}
\end{figure}

Figure~\ref{fig:summary} summarizes the signal-to-noise ratio (SNR) and DFA exponent $\alpha$ as functions of $\Delta T$. For the slot-car dataset, increasing $\Delta T$ improves SNR and drives $\alpha$ toward the random-noise limit, with both metrics asymptotically converging. Although selecting the largest possible $\Delta T$ is trivial for such a simple dataset—where BA noise is evidently random—this reasoning does not generally hold for complex scenes, where signal and noise cannot be visually separated and long-range noise correlations may exist.

Selecting an optimal $\Delta T$ therefore requires balancing denoising quality and computational cost. Larger $\Delta T$ values increase filter latency and complexity, as each event must be compared with others within the time window; a naive BA filter scales as $O(N^2)$. While more efficient implementations exist, they introduce additional trade-offs. In complex scenarios, where visual inspection is insufficient, statistical characterization of noise via DFA provides an objective criterion for selecting $\Delta T$.

\subsection{Complex dataset}

\begin{figure}[h]
\centering
\begin{subfigure}{0.24\textwidth}
\includegraphics[width=\textwidth]{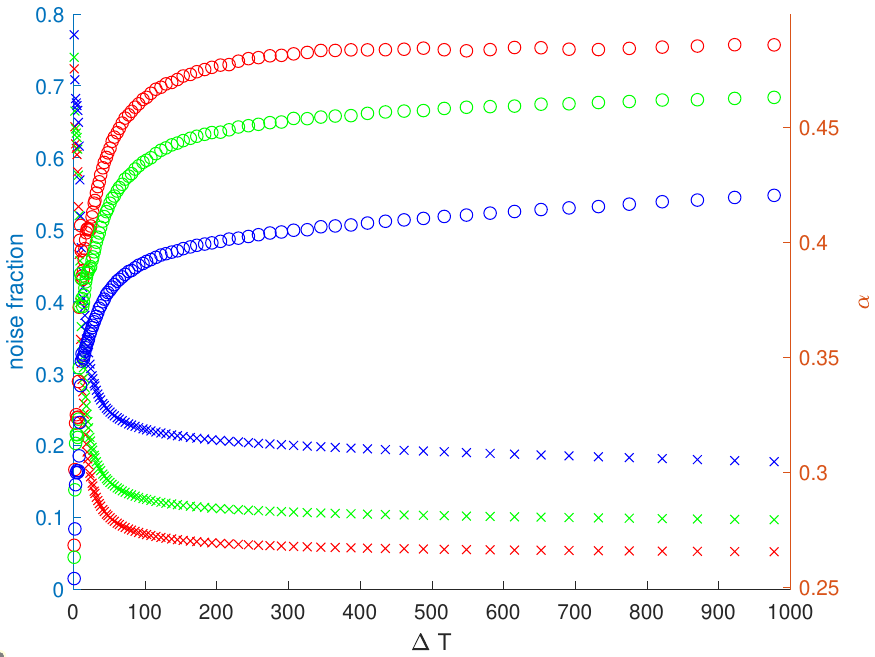} 
\subcaption{Spinner}
\label{fig:spinner}
\end{subfigure}
\begin{subfigure}{0.24\textwidth}
\includegraphics[width=\textwidth]{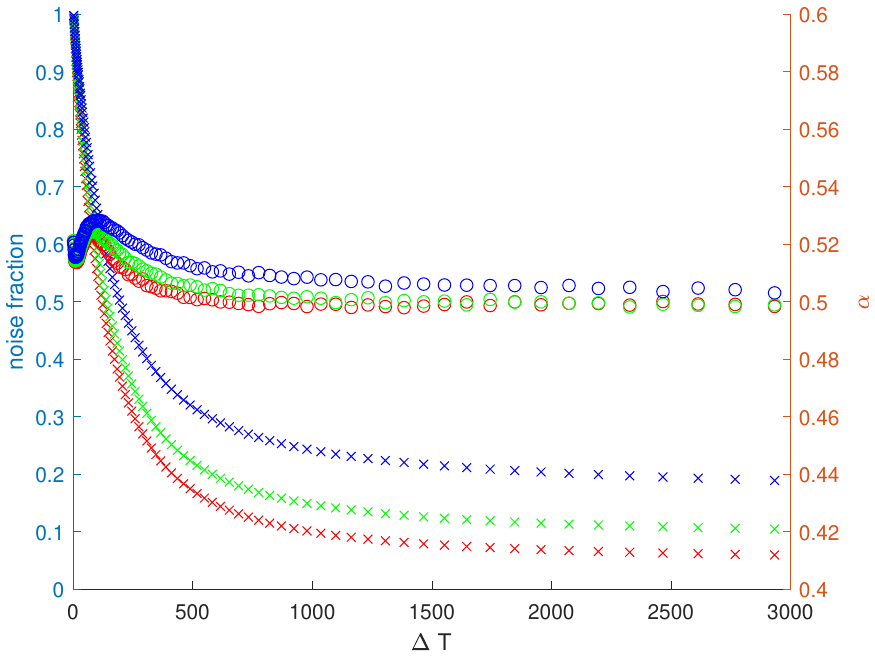}
\subcaption{Hand spinner}
\label{fig:hand_spinner}
\end{subfigure}
\begin{subfigure}{0.24\textwidth}
\includegraphics[width=\textwidth]{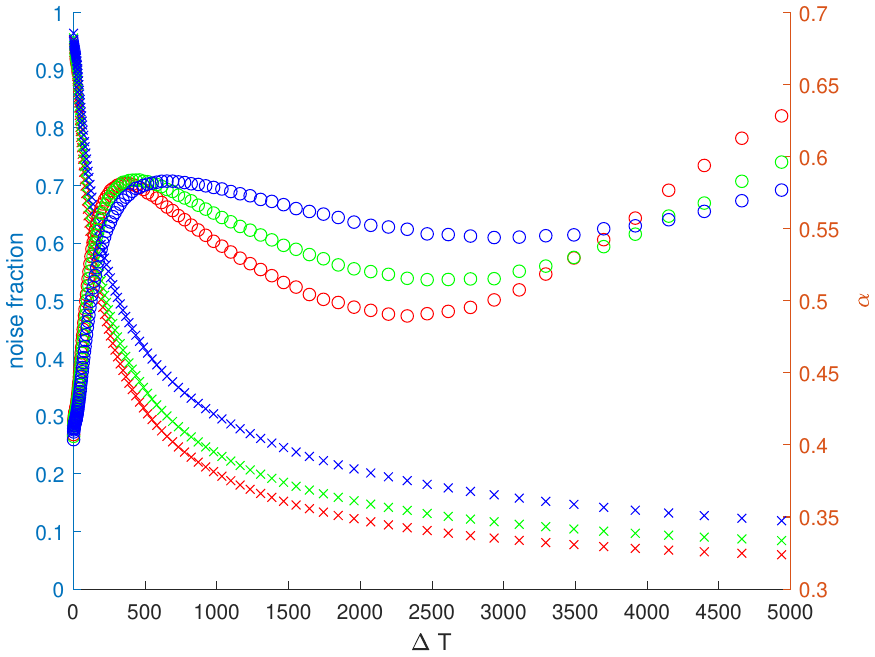}
\subcaption{Monitoring}
\label{fig:monitoring}
\end{subfigure}
\begin{subfigure}{0.24\textwidth}
\includegraphics[width=\textwidth]{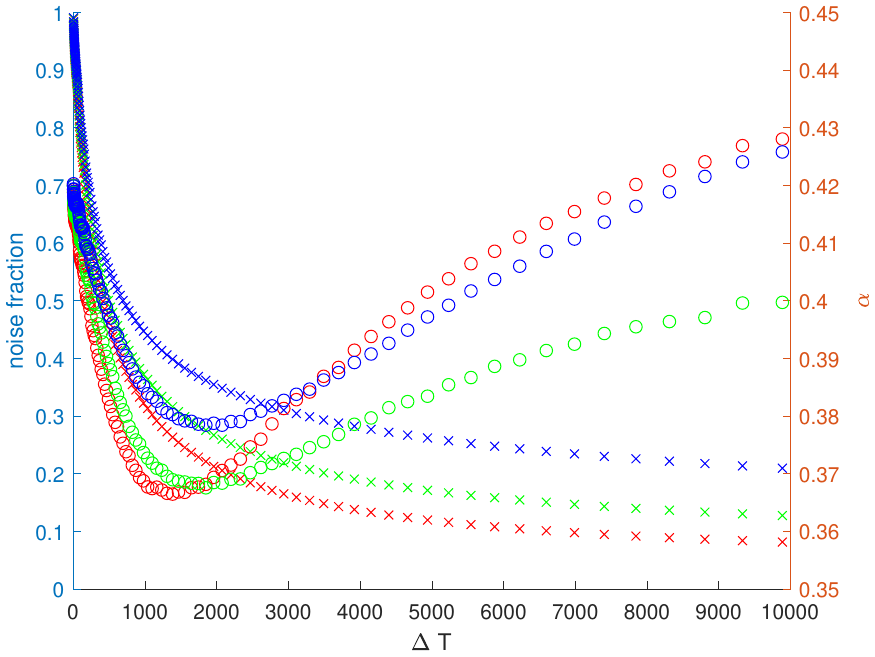}
\subcaption{Pedestrians}
\label{fig:pedestrians}
\end{subfigure}
\caption{ Noise fraction (left $y$-axis, circle marker) and DFA $\alpha$ noise (right $y$-axis, cross marker)  for complex PROPHESEE scenes injected by 0.05(red), 0.1(green), and 0.2(blue) BA noise.}
\end{figure}

\bluecolor{While arbitrary noisy complex scenes may require dedicated analysis, since the DFA exponent $\alpha$ can reflect a mixture of dynamics from multiple objects, the availability of ground truth enables direct validation of the proposed DFA-based criterion. Figures~\ref{fig:spinner}–\ref{fig:pedestrians} demonstrate that, under proper BA filtering, the DFA exponent of the residual noise converges toward $\alpha = 1/2$, as expected for uncorrelated noise. Clean event streams from the PROPHESSE dataset were used, with controlled BA noise injection. As the injected noise fraction increases to 0.05 (red), 0.1 (green), and 0.2 (blue) (left $y$-axis, circle markers), the corresponding DFA noise exponent $\alpha$ (right $y$-axis, cross markers) consistently approaches $1/2$. This result provides direct experimental validation that the DFA exponent serves as an objective and reliable indicator of effective BA noise suppression.} \bluecolorr{Varying illumination conditions may inherently alter the global firing rate of signal events. However, because DFA measures the scaling behavior of detrended fluctuations rather than absolute event frequencies, the convergence of the noise exponent $\alpha \rightarrow 0.5$ remains robust across different lighting scenarios, provided the underlying BA noise remains uncorrelated.}

\section{Conclusion}


Detrended Fluctuation Analysis (DFA) was introduced as an objective tool for characterizing the quality of denoised DVS data in the absence of ground truth. The proposed criterion is derived from the statistical properties of filtered background activity (BA) noise extracted from the DVS event stream. DFA was adapted to event-based data and validated on widely used datasets using a simple BA filter. The results demonstrate that DFA scaling exponents reliably capture residual signal components present in filtered BA noise, providing a quantitative indicator of denoising effectiveness. 

\bluecolorr{It is worth noting that spatial variations in DVS pixel sensitivities can lead to localized differences in event-generation rates, potentially causing slight deviations from the nominal baseline $\alpha$ value. Consequently, hardware-level threshold mismatches may affect the validity of the assumed random-noise criteria. Investigating these effects constitutes an important direction for future work, as it could enable more accurate sensor calibration and parameter selection.}

Future work will extend the analysis to more advanced BA filters, such as y-noise and k-noise, and to more complex datasets, including scenes with independently moving objects (IMO), where each object generates events with distinct statistical characteristics. Additional directions include studying the relationship between DFA exponents and Poisson statistics of inter-event intervals, as well as exploring alternative formulations in which the random variable is the number of events within a fixed time window $\Delta t_{\text{ref}}$, rather than inter-event time differences $x = t_i - t_{i-1}$.

\section{Acknowledgment}
This project has received funding from the European Union’s Horizon 2020 Research and Innovation Programme under Grant Agreement No 739578 and the Government of the Republic of Cyprus through the Deputy Ministry of Research, Innovation and Digital Policy.

\bibliographystyle{IEEEtran}


\begin{thebibliography}{10}
\providecommand{\url}[1]{#1}
\csname url@samestyle\endcsname
\providecommand{\newblock}{\relax}
\providecommand{\bibinfo}[2]{#2}
\providecommand{\BIBentrySTDinterwordspacing}{\spaceskip=0pt\relax}
\providecommand{\BIBentryALTinterwordstretchfactor}{4}
\providecommand{\BIBentryALTinterwordspacing}{\spaceskip=\fontdimen2\font plus
\BIBentryALTinterwordstretchfactor\fontdimen3\font minus \fontdimen4\font\relax}
\providecommand{\BIBforeignlanguage}[2]{{%
\expandafter\ifx\csname l@#1\endcsname\relax
\typeout{** WARNING: IEEEtran.bst: No hyphenation pattern has been}%
\typeout{** loaded for the language `#1'. Using the pattern for}%
\typeout{** the default language instead.}%
\else
\language=\csname l@#1\endcsname
\fi
#2}}
\providecommand{\BIBdecl}{\relax}
\BIBdecl

\bibitem{DVSsurvey2020}
G.~Gallego, T.~Delbrück, G.~Orchard, C.~Bartolozzi, B.~Taba, A.~Censi, S.~Leutenegger, A.~J. Davison, J.~Conradt, K.~Daniilidis, and D.~Scaramuzza, ``Event-based vision: A survey,'' \emph{IEEE Transactions on Pattern Analysis and Machine Intelligence}, vol.~44, no.~1, pp. 154--180, 2022.

\bibitem{Gallego2019FocusIA}
G.~Gallego, M.~Gehrig, and D.~Scaramuzza, ``Focus is all you need: Loss functions for event-based vision,'' \emph{2019 IEEE/CVF Conference on Computer Vision and Pattern Recognition (CVPR)}, pp. 12\,272--12\,281, 2019.

\bibitem{Muglikar2021HowTC}
M.~Muglikar, M.~Gehrig, D.~Gehrig, and D.~Scaramuzza, ``How to calibrate your event camera,'' \emph{2021 IEEE/CVF Conference on Computer Vision and Pattern Recognition Workshops (CVPRW)}, pp. 1403--1409, 2021.

\bibitem{Shiba2022SecretsOE}
S.~Shiba, Y.~Aoki, and G.~Gallego, ``Secrets of event-based optical flow,'' in \emph{European Conference on Computer Vision}, 2022.

\bibitem{Graa2023OptimalBA}
R.~Graça, B.~Mcreynolds, and T.~Delbr{\"u}ck, ``Optimal biasing and physical limits of dvs event noise,'' \emph{ArXiv}, vol. abs/2304.04019, 2023.

\bibitem{Graa2023ShiningLO}
------, ``Shining light on the dvs pixel: A tutorial and discussion about biasing and optimization,'' \emph{ArXiv}, vol. abs/2304.04706, 2023.

\bibitem{Delbruck2008FramefreeDD}
\BIBentryALTinterwordspacing
T.~Delbruck, ``\BIBforeignlanguage{english}{Frame-free dynamic digital vision},'' in \emph{\BIBforeignlanguage{english}{International Symposium on Secure-Life Electronics}}, vol.~1, no.~1.\hskip 1em plus 0.5em minus 0.4em\relax University of Tokyo, March 2008, pp. 21--26, in: Proceedings of International Symposium on Secure-Life Electronics, Advanced Electronics for Quality Life and Society, Univ. of Tokyo, Mar. 6-7, 2008. [Online]. Available: \url{https://doi.org/10.5167/uzh-17620}
\BIBentrySTDinterwordspacing

\bibitem{Liu2015DesignOA}
H.~Liu, C.~Brandli, C.~Li, S.-C. Liu, and T.~Delbr{\"u}ck, ``Design of a spatiotemporal correlation filter for event-based sensors,'' \emph{2015 IEEE International Symposium on Circuits and Systems (ISCAS)}, pp. 722--725, 2015.

\bibitem{khodamoradi2018n}
A.~Khodamoradi and R.~Kastner, ``O(n)-space spatiotemporal filter for reducing noise in neuromorphic vision sensors,'' \emph{IEEE Transactions on Emerging Topics in Computing}, 2018.

\bibitem{guo2022low}
S.~Guo and T.~Delbruck, ``Low cost and latency event camera background activity denoising,'' \emph{IEEE Transactions on Pattern Analysis and Machine Intelligence}, 2022.

\bibitem{feng2020event}
Y.~Feng, H.~Lv, H.~Liu, Y.~Zhang, Y.~Xiao, and C.~Han, ``Event density based denoising method for dynamic vision sensor,'' \emph{Applied Sciences}, 2020.

\bibitem{lagorce2016hots}
X.~Lagorce, G.~Orchard, F.~Galluppi, B.~E. Shi, and R.~B. Benosman, ``Hots: a hierarchy of event-based time-surfaces for pattern recognition,'' \emph{IEEE transactions on pattern analysis and machine intelligence}, pp. 1346--1359, 2016.

\bibitem{IETSfilter}
R.~W. Baldwin, M.~Almatrafi, J.~R. Kaufman, V.~Asari, and K.~Hirakawa, ``Inceptive event time-surfaces for object classification using neuromorphic cameras,'' in \emph{Image Analysis and Recognition}, F.~Karray, A.~Campilho, and A.~Yu, Eds.\hskip 1em plus 0.5em minus 0.4em\relax Cham: Springer International Publishing, 2019, pp. 395--403.

\bibitem{9091226}
J.~Wu, C.~Ma, L.~Li, W.~Dong, and G.~Shi, ``Probabilistic undirected graph based denoising method for dynamic vision sensor,'' \emph{IEEE Transactions on Multimedia}, vol.~23, pp. 1148--1159, 2021.

\bibitem{baldwin2020eventEDnCNN}
R.~W. Baldwin, M.~Almatrafi, V.~Asari, and K.~Hirakawa, ``Event probability mask (epm) and event denoising convolutional neural network (edncnn) for neuromorphic cameras,'' 2020.

\bibitem{Fang2022AEDNet}
\BIBentryALTinterwordspacing
H.~Fang, J.~Wu, L.~Li, J.~Hou, W.~Dong, and G.~Shi, ``Aednet: Asynchronous event denoising with spatial-temporal correlation among irregular data,'' in \emph{Proceedings of the 30th ACM International Conference on Multimedia}, ser. MM '22.\hskip 1em plus 0.5em minus 0.4em\relax New York, NY, USA: Association for Computing Machinery, 2022, p. 1427–1435. [Online]. Available: \url{https://doi.org/10.1145/3503161.3548048}
\BIBentrySTDinterwordspacing

\bibitem{Gallego2018}
G.~Gallego, H.~Rebecq, and D.~Scaramuzza, ``A unifying contrast maximization framework for event cameras, with applications to motion, depth, and optical flow estimation,'' in \emph{2018 IEEE/CVF Conference on Computer Vision and Pattern Recognition}, 2018, pp. 3867--3876.

\bibitem{Xu2023DenoisingFor}
N.~Xu, L.~Wang, J.~Zhao, and Z.~Yao, ``Denoising for dynamic vision sensor based on augmented spatiotemporal correlation,'' \emph{IEEE Transactions on Circuits and Systems for Video Technology}, vol.~33, no.~9, pp. 4812--4824, 2023.

\bibitem{Yang2018TheSC}
C.~Yang, H.~Feng, Z.~hai Xu, Q.~Li, and Y.~ting Chen, ``The spatial correlation problem of noise in imaging deblurring and its solution,'' \emph{J. Vis. Commun. Image Represent.}, vol.~56, pp. 167--176, 2018.

\bibitem{Peng1994}
\BIBentryALTinterwordspacing
C.-K. Peng, S.~V. Buldyrev, S.~Havlin, M.~Simons, H.~E. Stanley, and A.~L. Goldberger, ``Mosaic organization of dna nucleotides,'' \emph{Phys. Rev. E}, vol.~49, pp. 1685--1689, Feb 1994. [Online]. Available: \url{https://link.aps.org/doi/10.1103/PhysRevE.49.1685}
\BIBentrySTDinterwordspacing

\bibitem{Rogers2020-jn}
B.~Rogers, D.~Giles, N.~Draper, O.~Hoos, and T.~Gronwald, ``A new detection method defining the aerobic threshold for endurance exercise and training prescription based on fractal correlation properties of heart rate variability,'' \emph{Front. Physiol.}, vol.~11, p. 596567, 2020.

\bibitem{Kataoka2021}
\BIBentryALTinterwordspacing
T.~Kataoka, T.~Miyaguchi, and T.~Akimoto, ``Detrended fluctuation analysis of earthquake data,'' \emph{Phys. Rev. Res.}, vol.~3, p. 033081, Jul 2021. [Online]. Available: \url{https://link.aps.org/doi/10.1103/PhysRevResearch.3.033081}
\BIBentrySTDinterwordspacing

\bibitem{Gou2018lightningflash}
\BIBentryALTinterwordspacing
X.~Gou, M.~Chen, and G.~Zhang, ``Time correlations of lightning flash sequences in thunderstorms revealed by fractal analysis,'' \emph{Journal of Geophysical Research: Atmospheres}, vol. 123, no.~2, pp. 1351--1362, 2018. [Online]. Available: \url{https://agupubs.onlinelibrary.wiley.com/doi/abs/10.1002/2017JD027206}
\BIBentrySTDinterwordspacing

\bibitem{Shrestha2021finance}
\BIBentryALTinterwordspacing
K.~Shrestha, ``Multifractal detrended fluctuation analysis of return on bitcoin*,'' \emph{International Review of Finance}, vol.~21, no.~1, pp. 312--323, 2021. [Online]. Available: \url{https://onlinelibrary.wiley.com/doi/abs/10.1111/irfi.12256}
\BIBentrySTDinterwordspacing

\bibitem{Ramirez2005}
\BIBentryALTinterwordspacing
J.~Alvarez-Ramirez, E.~Rodriguez, I.~Cervantes, and J.~{Carlos Echeverria}, ``Scaling properties of image textures: A detrending fluctuation analysis approach,'' \emph{Physica A: Statistical Mechanics and its Applications}, vol. 361, no.~2, pp. 677--698, 2006. [Online]. Available: \url{https://www.sciencedirect.com/science/article/pii/S0378437105007193}
\BIBentrySTDinterwordspacing

\bibitem{Hu2001}
\BIBentryALTinterwordspacing
K.~Hu, P.~C. Ivanov, Z.~Chen, P.~Carpena, and H.~Eugene~Stanley, ``Effect of trends on detrended fluctuation analysis,'' \emph{Physical Review E}, vol.~64, no.~1, Jun. 2001. [Online]. Available: \url{http://dx.doi.org/10.1103/PhysRevE.64.011114}
\BIBentrySTDinterwordspacing

\bibitem{Kantelhardt2001}
\BIBentryALTinterwordspacing
J.~W. Kantelhardt, E.~Koscielny-Bunde, H.~H. Rego, S.~Havlin, and A.~Bunde, ``Detecting long-range correlations with detrended fluctuation analysis,'' \emph{Physica A: Statistical Mechanics and its Applications}, vol. 295, no.~3, pp. 441--454, 2001. [Online]. Available: \url{https://www.sciencedirect.com/science/article/pii/S0378437101001443}
\BIBentrySTDinterwordspacing

\bibitem{Taqqu1995}
\BIBentryALTinterwordspacing
M.~S. Taqqu, V.~Teverovsky, and W.~Willinger, ``Estimators for long-range dependence: An empirical study,'' \emph{Fractals}, vol.~03, no.~04, pp. 785--798, 1995. [Online]. Available: \url{https://doi.org/10.1142/S0218348X95000692}
\BIBentrySTDinterwordspacing

\bibitem{Hoell2019Theoret}
\BIBentryALTinterwordspacing
M.~H\"oll, K.~Kiyono, and H.~Kantz, ``Theoretical foundation of detrending methods for fluctuation analysis such as detrended fluctuation analysis and detrending moving average,'' \emph{Phys. Rev. E}, vol.~99, p. 033305, Mar 2019. [Online]. Available: \url{https://link.aps.org/doi/10.1103/PhysRevE.99.033305}
\BIBentrySTDinterwordspacing

\end{thebibliography}

\end{document}